\documentclass[11pt,a4paper]{article}
\usepackage[hyperref]{main}
\usepackage{times}
\usepackage{latexsym}
\usepackage{makecell}

\usepackage{graphicx}
\usepackage{comment}
\usepackage{amssymb}
\usepackage{amsmath}
\usepackage{multirow}
\usepackage{booktabs}
\usepackage{wasysym}

\usepackage{algorithm}
\usepackage{algorithmic}

\usepackage{soul}

\usepackage{microtype}

\aclfinalcopy 
\def\aclpaperid{1908} 

\title{{P}redicting {C}linical {T}rial {R}esults by {I}mplicit {E}vidence {I}ntegration}

\author{
Qiao Jin$^{1}$\thanks{\ \ Work done during internship at Alibaba DAMO Academy.}\hspace{0.3em},
Chuanqi Tan$^{2}$,
Mosha Chen$^{2}$,
Xiaozhong Liu$^{3}$,
Songfang Huang$^{2}$ \\
$^{1}$School of Medicine, Tsinghua University \\
$^{2}$Alibaba Group \\
$^{3}$Indiana University Bloomington \\
\tt{jqa14@mails.tsinghua.edu.cn} \\
\tt{\{chuanqi.tcq,chenmosha.cms,songfang.hsf\}@alibaba-inc.com}\\
\tt{liu237@indiana.edu}
}

\date{}

\begin{document}

\maketitle

\begin{abstract}
Clinical trials provide essential guidance for practicing Evidence-Based Medicine, 
though often accompanying with unendurable costs and risks. 
To optimize the design of clinical trials, 
we introduce a novel Clinical Trial Result Prediction (CTRP) task. 
In the CTRP framework, 
a model takes a PICO-formatted clinical trial proposal with its background as input and predicts the result,
i.e. how the \textbf{I}ntervention group compares with the \textbf{C}omparison group in terms of the measured \textbf{O}utcome in the studied \textbf{P}opulation.
While structured clinical evidence is prohibitively expensive for manual collection, 
we exploit large-scale unstructured sentences from medical literature that implicitly contain PICOs and results as evidence. 
Specifically, 
we pre-train a model to predict the disentangled results from such implicit evidence and fine-tune the model with limited data on the downstream datasets. 
Experiments on the benchmark Evidence Integration dataset show that the proposed model outperforms the baselines by large margins, e.g., with a 10.7\% relative gain over BioBERT in macro-F1. 
Moreover, the performance improvement is also validated on another dataset composed of clinical trials related to COVID-19.
\end{abstract}

\section{Introduction}
\label{intro}
Shall COVID-19 patients be treated with hydroxychloroquine?
In the era of 
Evidence-Based Medicine (EBM, \citealt{sackett1997evidence}),
medical practice should be guided by well-designed and well-conducted clinical research,
such as randomized controlled trials. However, conducting clinical trials is expensive and time-consuming. 
Furthermore, inappropriately designed studies can be devastating in a pandemic:
a high-profile Remdesivir clinical trial fails to achieve statistically significant conclusions \citep{wang2020remdesivir},
partially because it does not attain the predetermined sample size when ``competing with" other inappropriately designed trials that are unlikely to succeed or not so urgent to test (e.g.: physical exercises and dietary treatments). Therefore, it is crucial to carefully design and evaluate clinical trials before conducting them. 

Proposing new clinical trials requires support from previous evidence in medical literature or practice.
For example, 
the World Health Organization (WHO) has launched a global megatrial,
Solidarity \citep{Solidarity},
to prioritize clinical resources by
recommending only four most promising therapies\footnote{Remdesivir, lopinavir/ritonavir, interferon beta-1a and chloroquine/hydroxychloroquine.}.
The rationale for this suggestion comes from the integration of evidence that they might be effective against coronaviruses or other related organisms in laboratory or clinical studies \citep{peymani2016effect, sheahan2017broad, morra2018clinical}.
However, manual integration of evidence is far from satisfying, 
as one study reports that about 86.2\% of clinical trials fail \citep{wong2019estimation}
and even some of the Solidarity therapies do not get expected results \citep{mehra2020hydroxychloroquine}.

\begin{figure*}
    \centering
    \includegraphics[width=0.92\linewidth]{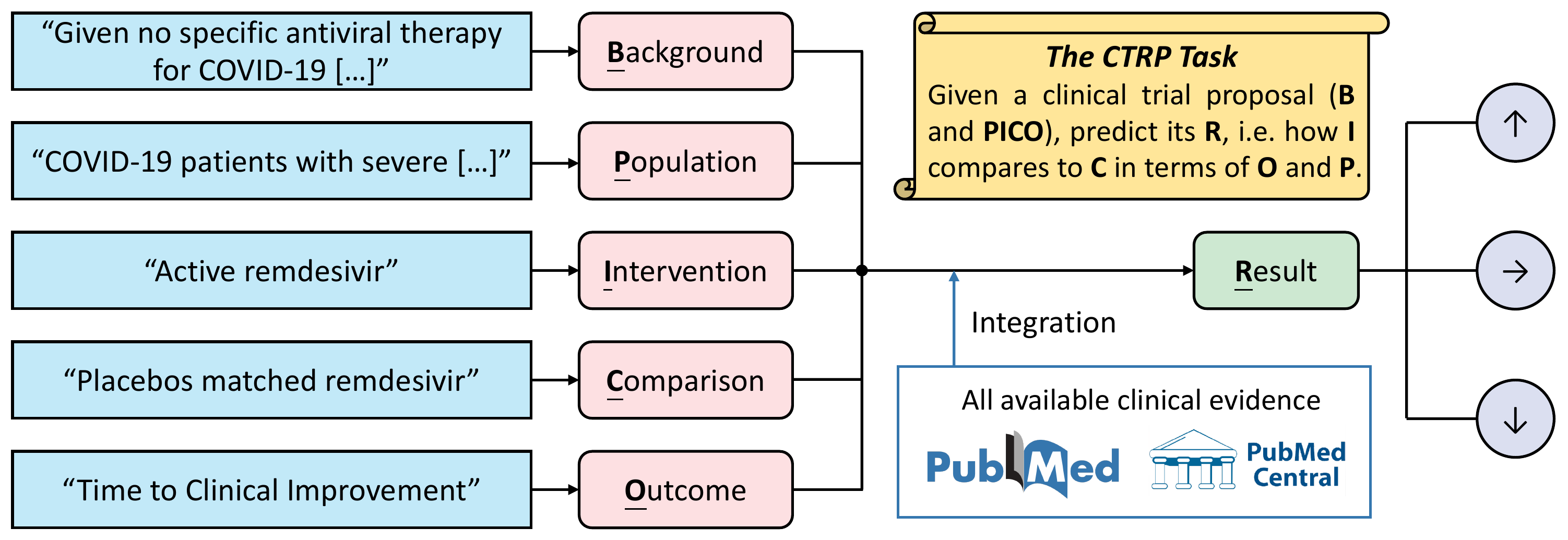}
    \caption{Architecture of the proposed Clinical Trial Result Prediction (CTRP) task.}
    \label{fig:task}
\end{figure*}

To assist clinical trial designing, we introduce a novel task:
Clinical Trial Result Prediction (CTRP), i.e. predicting the results of clinical trials without actually doing them (\S \ref{task}).
Figure \ref{fig:task} shows the architecture of the CTRP task.
We define the input to be a clinical trial proposal\footnote{The proposals need to be registered and approved before the clinical trials are conducted.},
which contains free-texts of a \textbf{P}opulation (e.g.: ``COVID-19 patients with severe symptoms"), an \textbf{I}ntervention (e.g.: ``Active remdesivir (i.v.)"), a \textbf{C}omparator (e.g.: ``Placebos matched remdesivir") and an \textbf{O}utcome (e.g.:``Time to clinical improvement"),
i.e. a PICO-formatted query \citep{huang2006evaluation},
and the background of the proposed trial. 
The output is the trial \textbf{R}esult, denoting 
how (higher, lower, or no difference) I compares to C in terms of O for P.

One particular challenge of this task is that evidence is entangled with other free-texts in the literature.
Prior works have explored \textit{explicit} methods for evidence integration through a pipeline of retrieval, extraction and inference on structured \{P,I,C,O,R\} evidence \citep{JMLR:v17:15-404, singh2017annotation, jin-szolovits-2018-pico, lee2018seed, nye-etal-2018-corpus, marshall-etal-2017-automating, lehman-etal-2019-inferring, deyoung2020evidence, zhang2020unlocking}.
However, they are limited in scale since getting domain-specific supervision for all clinical evidence is prohibitively expensive.

In this work,
we propose to \textit{implicitly} learn from such evidence by pre-training, 
instead of relying on \textit{explicit} evidence with purely supervised learning.
There are more than 30 million articles in PubMed\footnote{\url{https://pubmed.ncbi.nlm.nih.gov/}},
which stores almost all available medical evidence and thus is an ideal source for learning.
We collect 12 million sentences from PubMed abstracts and PubMed Central\footnote{\url{https://www.ncbi.nlm.nih.gov/pmc/}} (PMC) articles with comparative semantics,
which is commonly used to express clinical evidence (\S \ref{collection}).
P, I, C, O, and R are entangled with other free-texts in such sentences,
which we denote as implicit evidence.
Unlike previous efforts that seek to disentangle all of PICO and R,
we only disentangle R out of the implicit evidence using simple heuristics (\S \ref{disentanglement}).
For better learning the ordering function of I/C conditioned on P and O,
we also use adversarial examples generated by reversing both the entangled PICO and the R in the pre-training (\S \ref{adversarial}).
Then, we pre-train a transformer encoder \citep{vaswani2017attention} to predict the disentangled R from the implicit evidence,
which still contains PICO (\S \ref{clm}).
The model is named EBM-Net to reflect its utility for Evidence-Based Medicine.
Finally, 
we fine-tune the pre-trained EBM-Net on downstream datasets of the CTRP task (\S \ref{finetune}),
which are typically small in scale (\S \ref{experiments}).

To evaluate model performance,
we introduce a benchmark dataset, Evidence Integration (\S \ref{ei_dataset}), 
by re-purposing the \textit{evidence inference} dataset \citep{lehman-etal-2019-inferring, deyoung2020evidence}.
Experiments show that our pre-trained EBM-Net outperforms the baselines (\S \ref{baselines}) by large margins (\S \ref{main_results}).
Clustering analyses indicate that EBM-Net can effectively learn quantitative comparison results (\S \ref{discussions}).
In addition,
the EBM-Net model is further validated on a dataset composed of COVID-19 related clinical trials (\S \ref{covid}).

Our contribution is two-fold.
First, we propose a novel and meaningful task, CTRP, to predict clinical trial results before conducting them.
Second, unlike previous efforts that depend on structured data to understand the totality of clinical evidence,
we heuristically collect unstructured textual data, i.e. implicit evidence, and utilize large-scale pre-training to tackle the proposed CTRP task.
The datasets and codes are publicly available at \url{https://github.com/Alibaba-NLP/EBM-Net}.

\section{Related Works}
\paragraph{Predicting Clinical Trial Results:}
Most relevant works typically use only specific types or sources of information for prediction (e.g.: chemical structures \citep{gayvert2016data}, drug dosages or routes \citep{holford2000simulaton, holford2010clinical}).
\citet{gayvert2016data} predicts clinical trial results based on chemical properties of the candidate drugs.
Clinical trial simulation \citep{holford2000simulaton, holford2010clinical} applies pharmacological models to predict the results of a specific intervention with different procedural factors, 
such as doses and sampling intervals.
Some use closely related report information, e.g.: interim analyses \citep{broglio2014predicting} or phase II data for just phase II trials \citep{de2005predicting}.
Our task is (1) more generalizable,
since all potential PICO elements can be represented by free-texts and thus modeled in our work;
and (2) aimed at evaluating new clinical trial proposals.

\paragraph{Explicit Evidence Integration:} \label{explicit_evidence_integration} It depends on the existence of structured evidence, i.e.: \{P, I, C, O, R\} \citep{ijcai2019-899}.
Consequently,
collecting such explicit evidence is vital for further analyses,
and is also the objective for most relevant works:
Some seek to find relevant papers through retrieval \citep{lee2018seed};
many works are aimed at extracting PICO elements from published literature \citep{JMLR:v17:15-404, singh2017annotation, marshall-etal-2017-automating, jin-szolovits-2018-pico, nye-etal-2018-corpus,  zhang2020unlocking};
the \textit{evidence inference} task extracts R for a given ICO query using the corresponding clinical trial report \citep{lehman-etal-2019-inferring, deyoung2020evidence}.
However,
since getting expert annotations is expensive,
these works are typically limited in scale, 
with only thousands of labeled instances.
Few works have been done to utilize the automatically collected structured data for analyses.
In this paper, we adopt an end-to-end approach, 
where we use large-scale pre-training to \textit{implicitly} learn from free-text clinical evidence.

\section{The CTRP Task} \label{task}
The CTRP task is motivated to evaluate clinical trial proposals by predicting their results before actually conducting them,
as discussed in \S \ref{intro}.
Therefore, we formulate the task to take as input exactly the information required for proposing a new clinical trial:
free-texts of a background description and a PICO query to be investigated.
Formally, we denote the strings of the input background as $\texttt{B}$ and PICO elements as $\texttt{P}$, $\texttt{I}$, $\texttt{C}$, and $\texttt{O}$, respectively.
The task output is defined as one of the three possible comparison results: 
higher ($\uparrow$), no difference ($\rightarrow$), or lower ($\downarrow$) measurement $\texttt{O}$ in intervention group $\texttt{I}$ than in comparison group $\texttt{C}$ for population $\texttt{P}$.
We denote the result as $R$, and:
\[
R(\texttt{B},\texttt{P},\texttt{I},\texttt{C},\texttt{O}) = 
\begin{cases}
\uparrow & \texttt{O}(\texttt{I}) > \texttt{O}(\texttt{C})\ |\ \texttt{P}\  \\ 
\downarrow & \texttt{O}(\texttt{I}) < \texttt{O}(\texttt{C})\ |\ \texttt{P}\ \\ 
\rightarrow & \texttt{O}(\texttt{I}) \sim \texttt{O}(\texttt{C})\ |\ \texttt{P}\ \\ 
\end{cases}
\]
Main metrics include accuracy and 3-way macro-averaged F1.
We also use 2-way ($\uparrow$, $\downarrow$) macro-averaged F1 to evaluate human expectations (\S \ref{baselines}).

\begin{table*}[!t]
\centering
\small
\begin{tabular}{p{4.8cm}p{4.8cm}p{2.0cm}p{1.7cm}p{0.4cm}}
\toprule
$\texttt{E}_{ent}$ & $\texttt{E}_{dis}$ & $\texttt{R}$ & $r$ & $R$ \\
\midrule
``Our results also showed that serum TSH levels were \textcolor{red}{slightly higher} in the chloroquine group \textcolor{red}{than} in the placebo group." &
``Our results also showed that serum TSH levels were $\texttt{[MASK]}$ in the chloroquine group $\texttt{[MASK]}$ in the placebo group." & ``\textcolor{red}{slightly higher} ... \textcolor{red}{than}" & $\texttt{[HIGHER]}$ & $\uparrow$ \\
\midrule
``In conclusion, there is \textcolor{violet}{no difference between} IFN treatment \textcolor{violet}{and} supportive treatment for MERS patients in terms of mortality." &
``In conclusion, there is $\texttt{[MASK]}$ IFN treatment $\texttt{[MASK]}$ supportive treatment for MERS patients in terms of mortality." &
``\textcolor{violet}{no difference between} ... \textcolor{violet}{and}'' & $\texttt{[NODIFF]}$ & $\rightarrow$ \\
\midrule
``Levels of viral antigen staining in lung sections of GS-5734-treated animals were \textcolor{blue}{significantly lower as compared to} vehicle-treated animals." &
``Levels of viral antigen staining in lung sections of GS-5734-treated animals were $\texttt{[MASK]}$ vehicle-treated animals." &
``\textcolor{blue}{significantly lower as compared to}" &
$\texttt{[LOWER]}$ &
$\downarrow$ \\
\bottomrule
\end{tabular}
\caption{Several examples of implicit evidence. Red, violet and blue denote superiority, equality and inferiority.}
\label{tab:implicit_evidence}
\end{table*}

\section{Implicit Evidence Integration} \label{implicit_ei}

In this section, we introduce the Implicit Evidence Integration,
which is used to collect pre-training data for comparative language modeling (\S \ref{clm}).

Instead of collecting explicit evidence with structured $\{\texttt{B},\texttt{P},\texttt{I},\texttt{C},\texttt{O}-R\}$ information, 
we utilize a simple observation to collect evidence \textit{implicitly}:
clinical evidence is naturally expressed by comparisons,
e.g.: ``Blood oxygen is higher in the intervention group than in the placebo group".
Free-texts of $\texttt{P}$, $\texttt{I}$, $\texttt{C}$, $\texttt{O}$ and $\texttt{R}$ are entangled with other functional words that connect these elements in such comparative sentences,
where $\texttt{R}$ is a free-text version of the structured result $R$ (e.g.: $\texttt{R} = \text{``higher ... than"}$ translates into $R = \ \uparrow$).
We call these sentences entangled implicit evidence and denote them as 
$\texttt{E}_{ent} = \{\texttt{PICOR}\}$.
Then, we disentangle $\texttt{R}$ out of the $\texttt{E}_{ent}$ by heuristics, 
getting $\texttt{R}$ and the left $\texttt{E}_{dis} = \{\texttt{PICO}\}$.
We also include adversarial instances generated from the original ones.
Several examples are shown in Table \ref{tab:implicit_evidence}.

Details of implicit evidence collection, disentanglement, and adversarial data generation are introduced in \S\ref{collection}, \S\ref{disentanglement} and \S\ref{adversarial}, respectively.

\subsection{Collection of Implicit Evidence}
\label{collection}
We collect implicit evidence from PubMed abstracts and PMC articles\footnote{Articles in downstream experiment datasets are excluded.},
where most of the clinical evidence is published.
PubMed contains more than 30 million abstracts, 
and PMC has over 6 million full-length articles.
Each abstract is chunked into a background/method section and a result/conclusion section:
For the unstructured abstracts,
sentences before the first found implicit evidence are included in the background/method section.
For the semi-structured abstracts where each section is labeled with a section name,
the chunking is done by mapping the section name to either background/method or result/conclusion.

Sentences in abstract result/conclusion sections and main texts that express comparative semantics \cite{kennedy2004comparatives} are collected as implicit evidence.
They are identified by a pattern detection heuristic, similar to the keyword method described in \citet{jindal2006mining}: 
For expressions of superiority ($\uparrow$) and inferiority ($\downarrow$), 
we detect morpheme patterns of [\textit{more/less/-er ... than ...}].
For expression of equality ($\rightarrow$), 
we detect morpheme patterns of [\textit{similar ... to ...}] and [\textit{no difference ... between ... and ...}].
The background/method section serves as the corresponding $\texttt{B}$ for the collected implicit evidence.
These sentences are denoted as $\texttt{E}_{ent}$, which contain entangled PICO-R.

We have collected 11.8 million such sentences.
Among them, 2.4 million (20.2\%), 3.5 million (29.9\%) and 5.9 million (49.9\%) express inferiority, equality and superiority respectively.

\subsection{Disentanglement of Implicit Evidence}
\label{disentanglement}
To disentangle the free-text result $\texttt{R}$ from implicit evidence $\texttt{E}_{ent}$, 
we mask out the detected morphemes that express comparative semantics (e.g.: ``higher than") as well as other functional tokens that might be exploited by the model to predict the result (e.g.: p values).
This generates the masked out result $\texttt{R}$ and the left part $\texttt{E}_{dis}$ ($\{\texttt{PICO}\}$) from $\texttt{E}_{ent}$ ($\{\texttt{PICOR}\}$), i.e.: $\texttt{R} + \texttt{E}_{dis} = \texttt{E}_{ent}$.
$\texttt{R}$ is mostly a phrase with a central comparative adjective/adverb (e.g.: ``significantly smaller than") and can be directly mapped to $R$ ($\downarrow$ for the same example).

Nevertheless, $\texttt{R}$ contains richer information than the sole change direction because of the central adjective/adverb.
To utilize such information,
we map free-texts of $\texttt{R}$ to a finer-grained result label $r \in \mathcal{C}$ instead of the 3-way direction,
where $\mathcal{C} = \{\texttt{[POORER]}, \texttt{[LONGER]}, \texttt{[SLOWER]}, ...\}$ is a manually-curated vocabulary for such labels and $|\mathcal{C}| = 34$.
Each element can be mapped to its antonym in $\mathcal{C}$ by a reversing function $\text{Rev}$:
e.g.: $\text{Rev}(\texttt{[SMALLER]}) = \texttt{[GREATER]}$ and $\text{Rev}(\texttt{[NODIFF]}) = \texttt{[NODIFF]}$.
This enables us to generate adversarial examples used below.

\subsection{Adversarial Data Generation} \label{adversarial}
We generate adversarial examples from the original ones using a simple rule of ordering: 
if the result $r$ holds for the comparison $\texttt{I}/\texttt{C}$ conditioned on $\texttt{P}$ and $\texttt{O}$,
the reversed result $\text{Rev}(r)$ must hold for the reversed comparison $\texttt{C}/\texttt{I}$ on the same condition.
This is similar to generate adversarial examples for natural language inference task by logic rules \citep{minervini2018adversarially, wang2019if}.

However, 
Since $\texttt{E}_{dis}=\{\texttt{PICO}\}$ is only partially disentangled and $
\texttt{P}$, $\texttt{I}$, $\texttt{C}$, $\texttt{O}$ are still in their free-text forms,
we cannot explicitly reverse $\texttt{I}/\texttt{C}$ and generate such examples.
As an alternative, 
we reverse the entire sentence order while keeping the word order between any two masked phrases in $\texttt{E}_{dis}$,
getting $\texttt{E}_{rev}$. For example, if:

$\texttt{E}_{dis} = $ ``[Levels of viral antigen staining in lung sections of GS-5734-treated animals were]$_{0}$ $\texttt{[MASK]}$ [vehicle-treated animals]$_{1}$."

\noindent and $r = \texttt{[LOWER]}$, then the reversed evidence is: 

$\texttt{E}_{rev} = $ ``[Vehicle-treated animals]$_{1}$ $\texttt{[MASK]}$ [levels of viral antigen staining in lung sections of GS-5734-treated animals were]$_{0}$." 

\noindent and $\text{Rev}(r) = \texttt{[HIGHER]}$.
This implicitly reverses the ordering direction of $\texttt{I}$ and $\texttt{C}$ without changing the $\texttt{P}$ and $\texttt{O}$.

\begin{figure*}
    \centering
    \includegraphics[width=\linewidth]{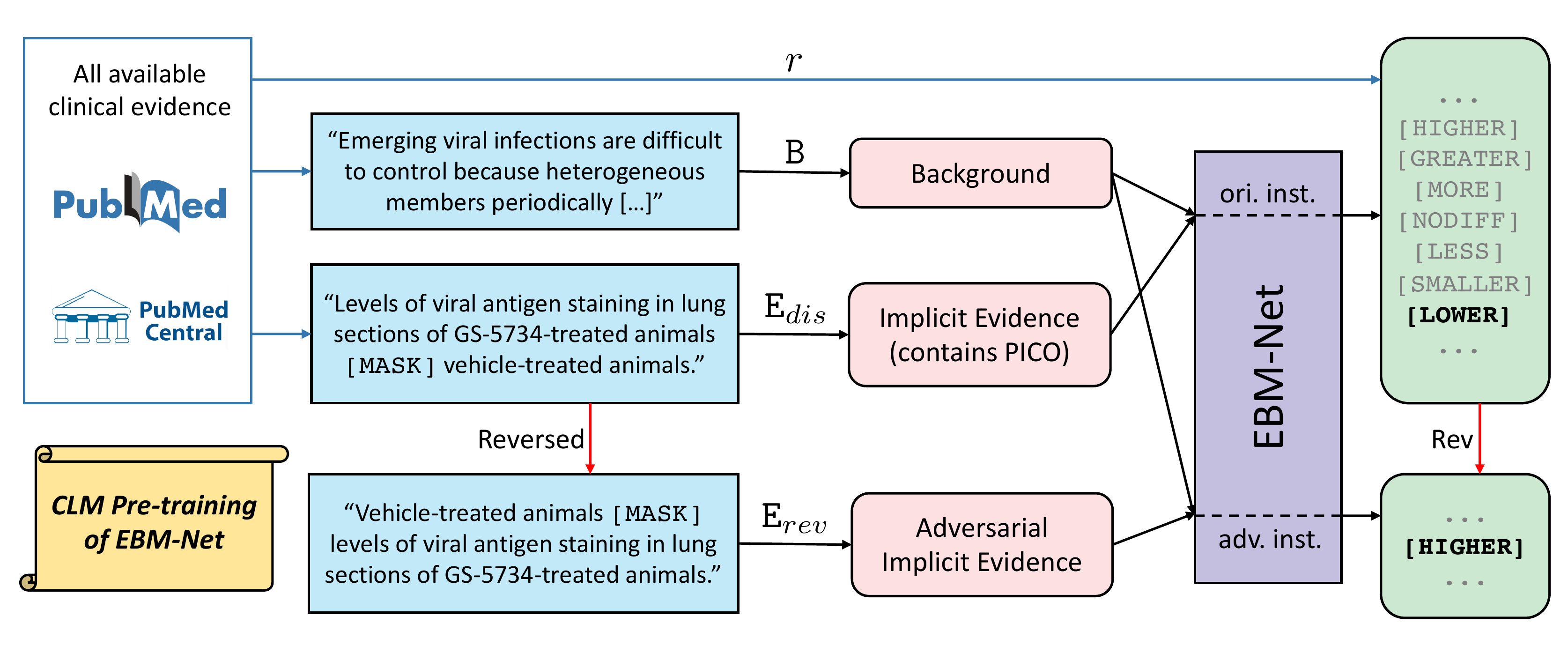}
    \caption{Architecture of CLM pre-training for EBM-Net. (ori.: original; adv.: adversarial; inst.: instance)}
    \label{fig:clm}
\end{figure*}

\section{EBM-Net}
\label{ebm_net}
We introduce the EBM-Net model in this section.
Similar to BERT \citep{devlin-etal-2019-bert}, 
EBM-Net is essentially a transformer encoder \citep{vaswani2017attention},
and follows the pre-training -- fine-tuning approach:
We pre-train EBM-Net by Comparative Language Modeling (CLM, \S \ref{clm}) that is designed to learn the conditional ordering function of $\texttt{I}$/$\texttt{C}$.
The pre-trained EBM-Net is fine-tuned to solve the CTRP task on downstream datasets (\S \ref{finetune}).

\subsection{Comparative Language Modeling} \label{clm}
We show the CLM architecture in Figure \ref{fig:clm}. 
CLM is adapted from the masked language modeling used in BERT \citep{devlin-etal-2019-bert},
but differentiates from it in that:
(1) EBM-Net masks out phrases $\texttt{R}$ that suggest comparative results and predicts a specific set of comparative labels $\mathcal{C}$;
(2) EBM-Net is also pre-trained on adversarial examples generated by comparison rules from the original examples.

During pre-training, 
EBM-Net takes as input the concatenation of background $\texttt{B}$ and the corresponding partially disentangled implicit evidence $\texttt{E}$,
i.e.: $\text{Input} = [\texttt{[CLS]}, \texttt{B}, \texttt{[SEP]}, \texttt{E}, \texttt{[SEP]}]$, 
where $\texttt{[CLS]}$ and $\texttt{[SEP]}$ are the special classification and separation tokens used in the original BERT and $\texttt{E} \in \{\texttt{E}_{dis}, \texttt{E}_{rev}\}$.
$\texttt{B}$ and $\texttt{E}$ are associated with two different segment types.
The special $\texttt{[MASK]}$ tokens are only used as placeholders for the masked out $\texttt{R}$.
$\texttt{[CLS]}$ hidden state of the EBM-Net is used to predict the CLM label $r$ with a linear layer followed by a softmax output unit:
\[\hat{r} = \text{SoftMax}(W_1h_{\texttt{[CLS]}} + b_1) \in [0,1]^{|\mathcal{C}|}\]
We minimize the cross-entropy between the estimated $\hat{r}$ and the empirical $r$ distribution.

At input-level, 
the adversarial examples only differ from their original examples in word orders between $\texttt{E}_{dis}$ and $\texttt{E}_{rev}$.
However, their labels are totally reversed from $r$ to $\text{Rev}(r)$.
By regularizing the model to learn such conditional ordering function,
CLM prevents the pre-trained model from learning unwanted and possibly biased co-occurrences between evident elements and their results.

\subsection{CTRP Fine-tuning} \label{finetune}
During fine-tuning,
EBM-Net takes as input the $[\texttt{[CLS]}, \texttt{B}, \texttt{[SEP]}, \texttt{E}_{exp}, \texttt{[SEP]}]$, 
where $\texttt{E}_{exp}$ denotes the explicit evidence in the downstream datasets of the proposed CTRP task.
For example, $\texttt{E}_{exp} = [\texttt{I}, \texttt{[SEP]}, \texttt{O}, \texttt{[SEP]}, \texttt{C}]$ on the Evidence Integration dataset (\S \ref{ei_dataset}).
The sequence of PICO elements in $\texttt{E}_{exp}$ can be tuned empirically.
EBM-Net learns from scratch another linear layer that maps from the predicted CLM label probabilities $\hat{r}$ to 3-way result label $R$ logits.
The final predictions are made by a softmax output unit:
\[\hat{R} = \text{SoftMax}(W_2\hat{r} + b_2) \in [0,1]^{3}\]
Cross-entropy between the estimated $\hat{R}$ and the empirical $R$ distribution is minimized in fine-tuning.

\subsection{Configuration}
The transformer weights of EBM-Net (L=12, H=768, A=12, \#Params=110M) are initialized with BioBERT \citep{lee2020biobert}, a variant of BERT that is also pre-trained on PubMed abstracts and PMC articles.
The maximum sequence lengths for $\texttt{B}$, $\texttt{E}_{dis}$, $\texttt{E}_{rev}$, $\texttt{E}_{exp}$ are 256, 128, 128, and 128, respectively.
We use Adam optimizer \citep{kingma2014adam} to minimize the cross-entropy losses.
EBM-Net is implemented using Huggingface's Transformers library \citep{Wolf2019HuggingFacesTS} in PyTorch \citep{NEURIPS2019_9015}.
Pre-training on 12M implicit evidence takes about 1k Tesla P100 GPU hours.

\section{Experiments} \label{experiments}
\subsection{The Evidence Integration Dataset} \label{ei_dataset}
The Evidence Integration dataset serves as a benchmark for our task.
We collect this dataset by re-purposing the \textit{evidence inference} dataset \citep{lehman-etal-2019-inferring, deyoung2020evidence},
which is essentially a machine reading comprehension task for extracting the structured result (i.e.: $R$) of a given structured ICO query\footnote{P is not included in the original dataset as the background of the trial report contains it.} from the corresponding clinical trial report article.
Since clinical trial reports already contain free-text result descriptions (i.e.: $\texttt{R}$) of the given ICO,
solving the original task does not require the integration of previous clinical evidence.
To test such capability for our proposed CTRP task, 
we remove the result/conclusion part and only keep the background/method part in the input clinical trial report.
34.6\% tokens of the original abstracts are removed on average and the remained are used as the clinical trial backgrounds.

Specifically, input of the Evidence Integration dataset includes free texts of ICO elements $\texttt{I}$, $\texttt{C}$ and $\texttt{O}$ which are the same as the original \textit{evidence inference} dataset, 
and their clinical trial backgrounds $\texttt{B}$.
The output is the comparison result $R$.
Following the original dataset split,
there are 8,164 instances for training,
1,002 for validation,
and 965 for test.

We also do experiments under the adversarial setting,
where adversarial examples generated by reversing both the $\texttt{I}$/$\texttt{C}$ order and the $R$ label (similar to \S \ref{clm})
are added.
This setting is used to test model robustness under adversarial attack.

\subsection{Compared Methods} \label{baselines}
We compare to a variety of methods, 
ranging from trivial ones like Random and Majority to the state-of-the-art BioBERT model.
Two major approaches in open-domain question answering (QA) are tested as well:
the knowledge base (KB) approach ({MeSH ontology}) and the text/retrieval approach ({Retrieval + Evidence Inference}),
since solving our task also requires reasoning over a large external corpus.
Finally, we introduce some ablation settings and the evaluation of human expectations.

\paragraph{Random:} 
we report the expected performance of randomly predicting the result for each instance.

\paragraph{Majority:} 
we report the performance of predicting the majority class ($\rightarrow$) for all test instances.n

\paragraph{Bag-of-Words + Logistic Regression:}
we concatenate the TF-IDF weighted bag-of-word vectors of $\texttt{B}$, $\texttt{P}$, $\texttt{I}$, $\texttt{C}$ and $\texttt{O}$ as features and use logistic regression for learning.

\paragraph{MeSH Ontology:} \label{ontology}
Since no external KB is available for our task,
we use the training set as an internal alternative:
we map the $\texttt{I}$, $\texttt{C}$ and $\texttt{O}$ of the test instances to terms in the Medical Subject Headings (MeSH)\footnote{\url{https://www.nlm.nih.gov/mesh}} ontology by string matching.
MeSH is a controlled and hierarchically-organized vocabulary for describing biomedical topics.
Then, we find their nearest labeled instances in the training set,
where the distance is defined by: 
\[d(i, j) = \sum_{e \in \{\texttt{I}, \texttt{C}, \texttt{O}\}} \min \text{TreeDist}(m_{i}^{e}, m_{j}^{e}) \]
$m_{i}^{e}$ and $m_{j}^{e}$ are MeSH terms identified in ICO element $e$ of instance $i$ and $j$, respectively. TreeDist is defined as the number of edges between two nodes on the MeSH tree.
The majority label of the nearest training instances is used as the prediction.

\paragraph{Retrieval + Evidence Inference:} \label{retrieval_baseline}
State-of-the-art method on the \textit{evidence inference} dataset \citep{deyoung2020evidence} is a pipeline based on SciBERT \citep{beltagy2019scibert}:
(1) find the exact evidence sentences in the clinical trial report for the given ICO query,
using a scoring function derived from a fine-tuned SciBERT;
and (2) predict the result $R$ based on the found evidence sentences and the given ICO query by fine-tuning another SciBERT.

Our task needs an additional retrieval step to find relevant documents that might contain useful results of similar trials,
as the input trial background does not contain the result information for the given ICO query.
Documents are retrieved from the entire PubMed and PMC using a TF-IDF matching between their indexed MeSH terms and the MeSH terms identified in the ICO queries.
We then apply the pipeline described above on the retrieved documents.
This baseline is similar to but more domain-specific than BERTserini \citep{yang-etal-2019-end}.

\paragraph{BioBERT:}
For this setting,
we feed BioBERT with similar input to EBM-Net as is described in \S\ref{ebm_net} and fine-tune it to predict the $R$ using its special $\texttt{[CLS]}$ hidden state.

\paragraph{Ablations:}
We conduct two sets of ablation experiments with EBM-Net:
(1) Pre-training level, 
where we exclude the adversarial examples in pre-training,
to analyze the utility of CLM against traditional LM.
(2) Input level, 
where we exclude different input elements ($\texttt{B}$, $\texttt{I}$, $\texttt{C}$, $\texttt{O}$) to study their relative importance.

\begin{table*}
    \small
    \centering
    \begin{tabular}{lccccccc}
        \toprule
        \multirow{2}{*}{\textbf{Model}} & 
        \multicolumn{3}{c}{\textbf{Standard Evidence Integration}} &
        \multicolumn{3}{c}{\textbf{Adversarial Evidence Integration}} & \multirow{2}{*}{$|\Delta|$}\\
        \cmidrule{2-7}
        & {Accuracy} & {F1} (3-way) & {F1} (2-way) & {Accuracy} & {F1} (3-way) & {F1} (2-way)  \\
        \midrule
        Majority ($\rightarrow$) & 41.76 & 19.64 & -- & 41.76 & 19.64 & -- & -- \\
        Random (expected) & 33.33 & 32.77 & 30.62 & 33.33 & 32.77 & 30.62  & -- \\
        \midrule
        BoW + Logistic Regression & 43.73 & 41.04 & 35.84 & 41.97 & 39.87 & 34.01 & 4.0  \\
        MeSH Ontology & 38.55 & 36.33 & 31.01 & 34.46 & 33.19 & 34.77 & 10.6 \\
        \makecell{Retrieval + Evidence Inference \\\citep{deyoung2020evidence}} & 50.57 & 49.91 & 48.30 & 50.62 & 50.13 & 48.46 & 0.0 \\
        BioBERT \citep{lee2020biobert} & 55.96 & 54.33 & 51.98 & 53.11 & 52.84 & 51.59 & 5.1 \\
        \midrule
        EBM-Net (ours) & \textbf{61.35} & \textbf{60.15} & \textbf{59.42} & \textbf{59.59} & \textbf{59.36} & \textbf{58.67} & 2.7 \\
        \hspace{3mm} w/o adversarial pre-training & 60.73 & 59.04 & 58.52 & 58.91 & 58.81 & 58.34 & 3.0\\
        \hspace{3mm} w/o $\texttt{B}$ (background) & 55.65 & 54.31 & 52.48 & 53.83 & 53.32 & 51.26 & 3.3 \\
        \hspace{3mm} w/o $\texttt{I}$ (intervention) & 59.59 & 58.59 & 58.08 & 57.30 & 56.74 & 54.87 & 3.8 \\
        \hspace{3mm} w/o $\texttt{C}$ (comparison) & 57.72 & 56.77 & 56.15 & 57.51 & 57.10 & 55.47 & 0.4 \\
        \hspace{3mm} w/o $\texttt{O}$ (outcome) & 48.91 & 44.88 & 39.57 & 47.31 & 46.40 & 43.66 & 3.3\\
        \midrule
        Human Expectations & 56.79 & -- & 68.86 & 56.79 & -- & 68.86 & -- \\
        \bottomrule
    \end{tabular}
    \caption{Main results on the benchmark Evidence Integration dataset. $|\Delta|$ denotes the absolute value of relative accuracy decrease from the standard to the adversarial setting. All numbers are percentages. (w/o: without)}
    \label{tab:main_result}
\end{table*}

\paragraph{Human Expectations:}
We define the expected result ($R_e$) of a clinical trial (e.g.: $R_e = \ \downarrow$ for $\texttt{O} = \text{``mortality rate"}$) as the Human Expectation (HE),
which is the underlying motivation for conducting the corresponding trial.
Generally, 
$R_e \in \{\uparrow, \downarrow\}$ since significant results are expected.
To make fair comparisons,
we use the 2-way macro-average F1:
$\text{F1 (2-way)} = (\text{F1}(\uparrow) + \text{F1}(\downarrow)) / 2$ as a main metric for evaluations of HE.
HE performance is an over-estimation of human performance:
main biases are due to the shift of input trial distribution from the targeted proposal stage to the actual report stage,
which contains fewer trials with unexpected results.

\subsection{Main Results}  \label{main_results}
Table \ref{tab:main_result} shows the main results on the Evidence Integration dataset, 
where accuracy and F1 (3-way) are used to compare model performance and F1 (2-way) is used for evaluating human expectations.

Results show that EBM-Net outperforms other baselines by large margins in both standard and adversarial settings.
While being the strongest baseline,
BioBERT is 10.7\% relatively lower in macro-F1 (54.33\% v.s. 60.15\%) and 9.6\% relatively lower in accuracy (55.96\% v.s. 61.35\%) than EBM-Net.
The open-domain QA baselines perform even worse:
for the MeSH Ontology method,
the internal KB of only 8k entries is far from complete;
for the Retrieval + Evidence Inference method, 
the PICO queries are so specific that no exactly relevant evidence can be found in other trials and retrieving only a few trials has limited utilities.

We use $|\Delta|$, the absolute value of relative accuracy decrease to measure model robustness under adversarial attacks.
The higher the $|\Delta|$, the more vulnerable a model is.
BioBERT has about twice as much (5.1\% v.s. 2.7\%) $|\Delta|$ in the adversarial setting as EBM-Net does.
It suggests that EBM-Net is more robust to adversarial attacks,
which is a vital property for healthcare applications.
EBM-Net without adversarial pre-training is less robust than EBM-Net as well (3.0\% v.s. 2.7\%),
but not as vulnerable as BioBERT,
indicating that robustness can be learned by pre-training with original implicit evidence to some extent and further consolidated by the adversarial evidence.

Unsurprisingly, EBM-Net with full input consistently outperforms all input-level ablations.
Among them,
$\texttt{O}$ is the most important input element as the performance decreases dramatically on its ablation.
This is expected as $\texttt{O}$ is the standard of comparisons.
$\texttt{B}$ is the second most important element,
since $\texttt{B}$ contains methodological details of how the clinical trials will be conducted,
which is also vital for result prediction.
The performance does not decrease as much without $\texttt{I}$ or $\texttt{C}$, 
since there is redundant information of them in $\texttt{B}$.

On the one hand,
the accuracy of EBM-Net surpasses that of HE,
mainly because the latter is practically a 2-way classifier.
On the other hand,
HE outperforms EBM-Net in terms of 2-way F1,
but is still unsatisfying (68.86\%).
This suggests that the proposed CTRP task is hard and there is still room for further improvements.

\subsection{Discussions} \label{discussions}
\begin{figure}
    \centering
    \includegraphics[width=\linewidth]{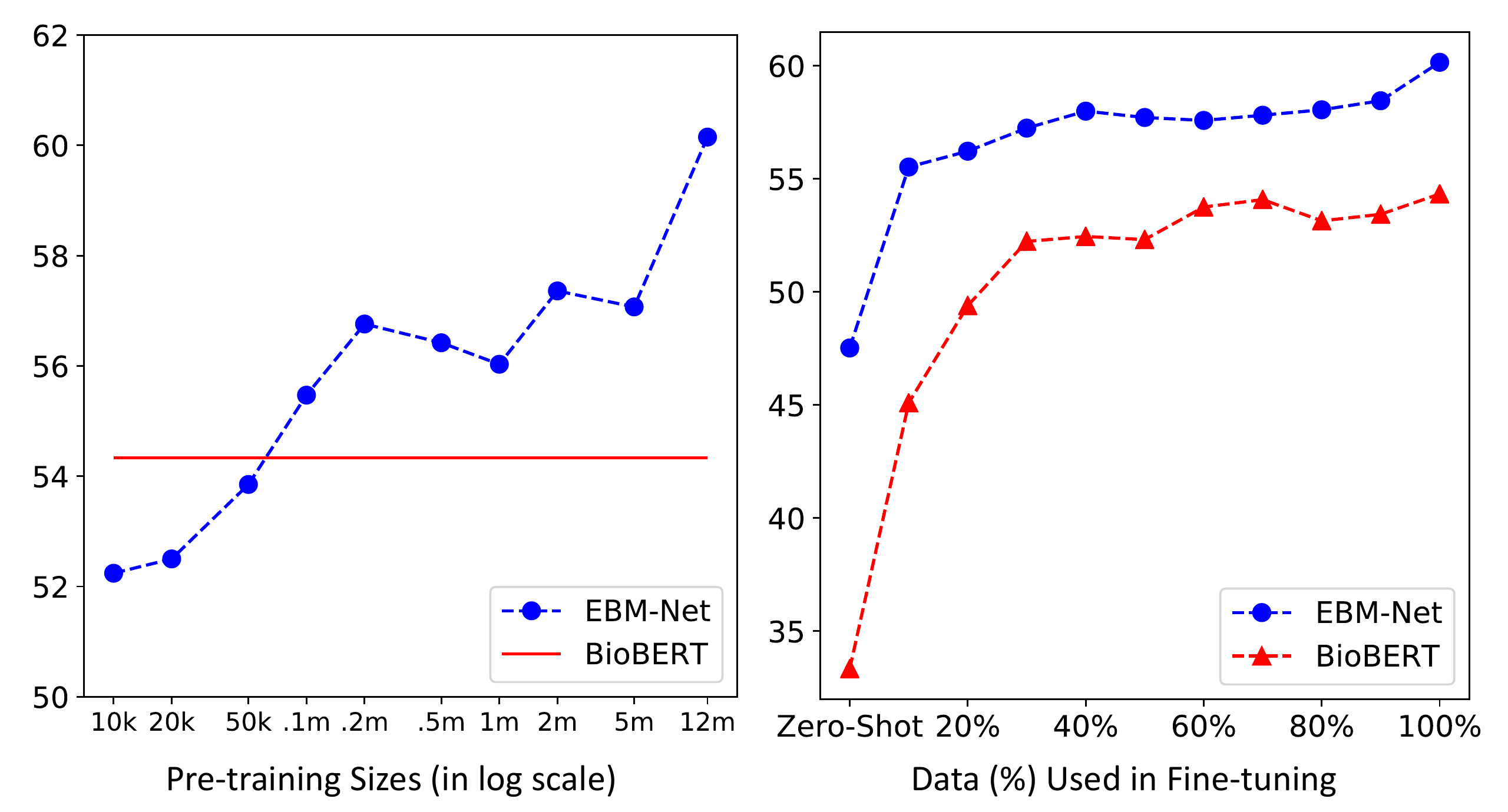}
    \caption{\textbf{Left:} EBM-Net 3-way macro-F1 v.s. pre-training sizes compared to BioBERT; \textbf{Right:} EBM-Net and BioBERT 3-way macro-F1 v.s. fine-tuning sizes.}
    \label{fig:sizes}
\end{figure}
We study how different numbers of pre-training and fine-tuning instances influence the EBM-Net performance,
in comparison to the BioBERT.
Figure \ref{fig:sizes} shows the results:
(\textbf{Left}) The final performance of EBM-Net improves log-linearly as the pre-training dataset size increases,
suggesting that there can be further improvements if more data is collected for pre-training but the marginal utility might be small.
EBM-Net surpasses BioBERT when pre-trained by about 50k to 100k instances of implicit evidence,
which are 5 to 10 times as many as the fine-tuning instances.
(\textbf{Right}) EBM-Net is more robust in a few-shot learning setting:
using only 10\% of the training data, 
EBM-Net outperforms BioBERT fine-tuned with 100\% of the training data.
From zero-shot\footnote{Zero-shot performance of BioBERT is defined as the expected results from random predictions.} to using all the training data,
EBM-Net improves only by 26.6\% relative F1 (from 47.52\% to 60.15\%) while BioBERT improves largely by 60.0\% relative F1 (from 32.77\% to 54.33\%).

\begin{figure}
    \centering
    \includegraphics[width=0.92\linewidth]{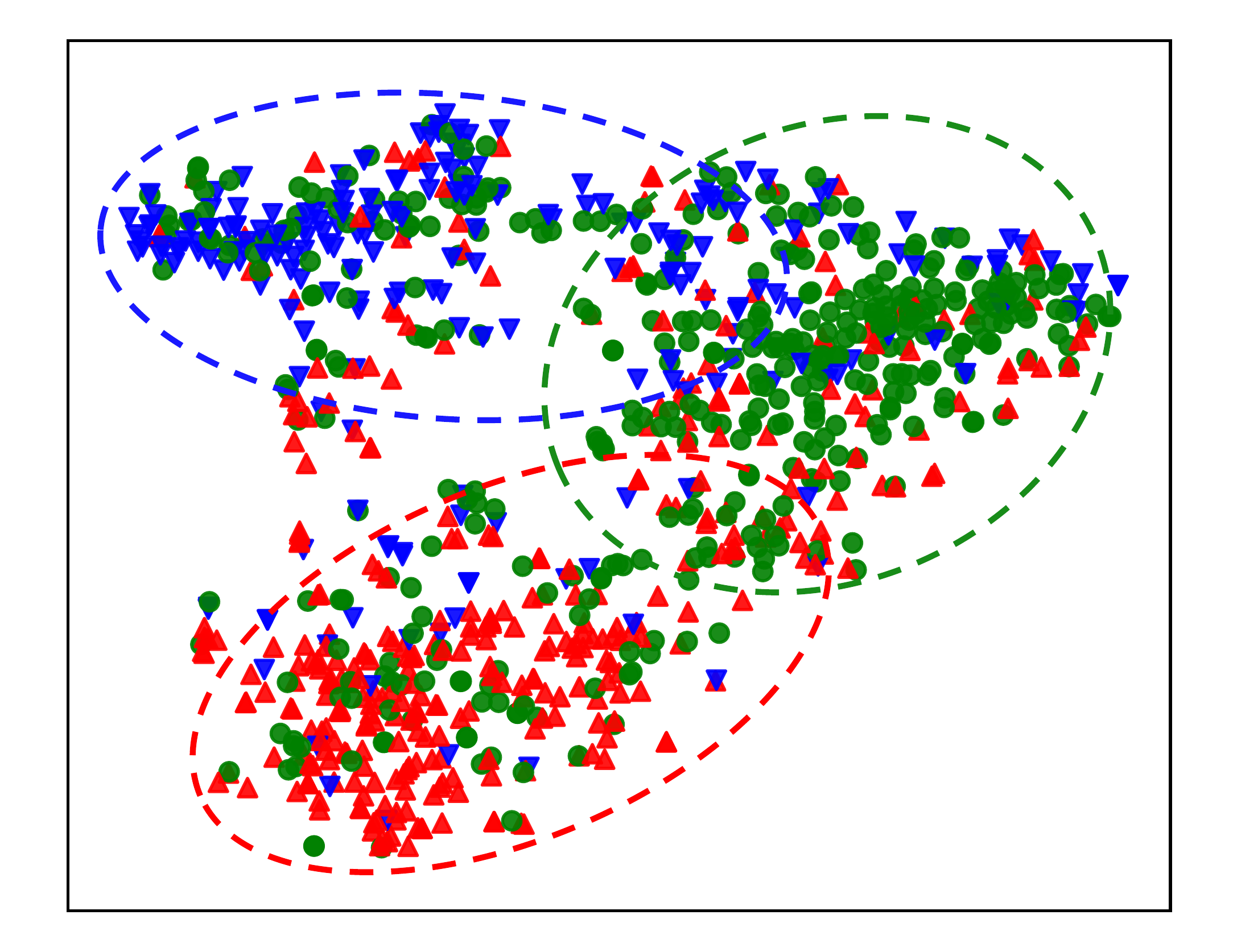}
    \caption{T-SNE visualizations of EBM-Net representations of Evidence Integration test set instances. Red colored $\blacktriangle$, blue colored $\blacktriangledown$, and green colored $\CIRCLE$ refer to the corresponding $R$ equaling $\uparrow$, $\downarrow$ and $\rightarrow$, respectively.}
    \label{fig:cluster}
\end{figure}

We use t-SNE \citep{maaten2008visualizing} to visualize the test instance representations derived from EBM-Net $\texttt{[CLS]}$ hidden state in Figure \ref{fig:cluster}.
It shows that EBM-Net effectively learns the relationships between comparative results:
the points cluster into three results ($\uparrow$, $\downarrow$, $\rightarrow$).
While there is a clear boundary between the $\downarrow$ cluster (dashed-blue circle) and the $\uparrow$ cluster (dashed-red circle),
the boundaries between the $\rightarrow$ cluster (dashed-black circle) and the other two are relatively vague.
It suggests that the learnt manifold follows a quantitatively continuous ``$\downarrow$ -- $\rightarrow$ -- $\uparrow$" pattern.

Out of the 373 mistakes EBM-Net makes on the test set,
significantly less (11.8\%, p$<$0.001 by permutation test) predictions are opposite to the ground-truth (e.g.: predicting $\uparrow$ when the label is $\downarrow$),
also suggesting that EBM-Net effectively learn the relationship between comparison results.
In addition, we notice that there is a considerable proportion of instances whose results are not predictable without their exact reports.
For example, 
some $\texttt{I}$ and $\texttt{C}$ differ only quantitatively,
e.g.: ``4\% lidocaine" and ``2\% lidocaine",
and modeling such differences is beyond the scope of our task.

\subsection{Validation on COVID-19 Clinical Trials} \label{covid}
For analyzing COVID-19 related clinical trials, 
we further pre-train EBM-Net on the CORD-19 dataset \citep{wang2020cord}\footnote{The 05/12/2020 version.},
also using the comparative language modeling (\S\ref{clm}).
It leads to a COVID-19 specific EBM-Net that is used in this section.

We use leave-one-out validation to evaluate EBM-Net on the 22 completed clinical trials in COVID-evidence\footnote{\url{covid-evidence.org} (visited on 05/18/2020).}, 
which is an expert-curated database of available evidence on interventions for COVID-19.
Again, EBM-Net outperforms BioBERT by a large margin (59.1\% v.s. 50.0\% accuracy).
Expectedly, their 3-way F1 results (45.5\% v.s. 36.1\%) are close to those in the zero-shot learning setting since not many trials have finished.
Accuracy and 2-way F1 performance of HE are 54.5\% and 68.9\%,
and are close to those in Table \ref{tab:main_result}.
These further confirm the performance improvement of EBM-Net and the difficulty of the CTRP task.

\section{Conclusions}
In this paper, we introduce a novel task, CTRP,
to predict clinical trial results without actually doing them.
Instead of using structured evidence that is prohibitively expensive to annotate,
we heuristically collect 12M unstructured sentences as implicit evidence, 
and use large-scale CLM pre-training to learn the conditional ordering function required for solving the CTRP task.
Our EBM-Net model
outperforms other strong baselines on the Evidence Integration dataset and is also validated on COVID-19 clinical trials.

\section*{Acknowledgement}
We would like to thank the anonymous reviewers of EMNLP 2020 for their constructive comments. We are also grateful for Ning Ding, Yuxuan Lai, Yijia Liu, Yao Fu, Kun Liu and Rui Wang for helpful discussions at Alibaba DAMO Academy.

\bibliography{main}

\begin{thebibliography}{38}
\expandafter\ifx\csname natexlab\endcsname\relax\def\natexlab#1{#1}\fi

\bibitem[{Beltagy et~al.(2019)Beltagy, Lo, and Cohan}]{beltagy2019scibert}
Iz~Beltagy, Kyle Lo, and Arman Cohan. 2019.
\newblock Scibert: A pretrained language model for scientific text.
\newblock In \emph{Proceedings of the 2019 Conference on Empirical Methods in
  Natural Language Processing and the 9th International Joint Conference on
  Natural Language Processing (EMNLP-IJCNLP)}, pages 3606--3611.

\bibitem[{Broglio et~al.(2014)Broglio, Stivers, and
  Berry}]{broglio2014predicting}
Kristine~R Broglio, David~N Stivers, and Donald~A Berry. 2014.
\newblock Predicting clinical trial results based on announcements of interim
  analyses.
\newblock \emph{Trials}, 15(1):73.

\bibitem[{De~Ridder(2005)}]{de2005predicting}
Filip De~Ridder. 2005.
\newblock Predicting the outcome of phase iii trials using phase ii data: a
  case study of clinical trial simulation in late stage drug development.
\newblock \emph{Basic \& clinical pharmacology \& toxicology}, 96(3):235--241.

\bibitem[{Devlin et~al.(2019)Devlin, Chang, Lee, and
  Toutanova}]{devlin-etal-2019-bert}
Jacob Devlin, Ming-Wei Chang, Kenton Lee, and Kristina Toutanova. 2019.
\newblock \href {https://doi.org/10.18653/v1/N19-1423} {{BERT}: Pre-training of
  deep bidirectional transformers for language understanding}.
\newblock In \emph{Proceedings of the 2019 Conference of the North {A}merican
  Chapter of the Association for Computational Linguistics: Human Language
  Technologies, Volume 1 (Long and Short Papers)}, pages 4171--4186,
  Minneapolis, Minnesota. Association for Computational Linguistics.

\bibitem[{DeYoung et~al.(2020)DeYoung, Lehman, Nye, Marshall, and
  Wallace}]{deyoung2020evidence}
Jay DeYoung, Eric Lehman, Ben Nye, Iain~J. Marshall, and Byron~C. Wallace.
  2020.
\newblock \href {http://arxiv.org/abs/2005.04177} {Evidence inference 2.0: More
  data, better models}.

\bibitem[{Gayvert et~al.(2016)Gayvert, Madhukar, and
  Elemento}]{gayvert2016data}
Kaitlyn~M Gayvert, Neel~S Madhukar, and Olivier Elemento. 2016.
\newblock A data-driven approach to predicting successes and failures of
  clinical trials.
\newblock \emph{Cell chemical biology}, 23(10):1294--1301.

\bibitem[{Holford et~al.(2010)Holford, Ma, and Ploeger}]{holford2010clinical}
N~Holford, SC~Ma, and BA~Ploeger. 2010.
\newblock Clinical trial simulation: a review.
\newblock \emph{Clinical Pharmacology \& Therapeutics}, 88(2):166--182.

\bibitem[{Holford et~al.(2000)Holford, Kimko, Monteleone, and
  Peck}]{holford2000simulaton}
N.~H.~G. Holford, H.~C. Kimko, J.~P.~R. Monteleone, and C.~C. Peck. 2000.
\newblock \href {https://doi.org/10.1146/annurev.pharmtox.40.1.209} {Simulation
  of clinical trials}.
\newblock \emph{Annual Review of Pharmacology and Toxicology}, 40(1):209--234.
\newblock PMID: 10836134.

\bibitem[{Huang et~al.(2006)Huang, Lin, and
  Demner-Fushman}]{huang2006evaluation}
Xiaoli Huang, Jimmy Lin, and Dina Demner-Fushman. 2006.
\newblock Evaluation of pico as a knowledge representation for clinical
  questions.
\newblock In \emph{AMIA annual symposium proceedings}, volume 2006, page 359.
  American Medical Informatics Association.

\bibitem[{Jin and Szolovits(2018)}]{jin-szolovits-2018-pico}
Di~Jin and Peter Szolovits. 2018.
\newblock \href {https://doi.org/10.18653/v1/W18-2308} {{PICO} element
  detection in medical text via long short-term memory neural networks}.
\newblock In \emph{Proceedings of the {B}io{NLP} 2018 workshop}, pages 67--75,
  Melbourne, Australia. Association for Computational Linguistics.

\bibitem[{Jindal and Liu(2006)}]{jindal2006mining}
Nitin Jindal and Bing Liu. 2006.
\newblock Mining comparative sentences and relations.
\newblock In \emph{Proceedings of the 21st National Conference on Artificial
  Intelligence - Volume 2}, AAAI’06, page 1331–1336. AAAI Press.

\bibitem[{Kennedy(2004)}]{kennedy2004comparatives}
Christopher Kennedy. 2004.
\newblock Comparatives, semantics of.
\newblock \emph{Concise Encyclopedia of Philosophy of Language and
  Linguistics}, pages 68--71.

\bibitem[{Kingma and Ba(2014)}]{kingma2014adam}
Diederik~P Kingma and Jimmy Ba. 2014.
\newblock Adam: A method for stochastic optimization.
\newblock \emph{arXiv preprint arXiv:1412.6980}.

\bibitem[{Lee and Sun(2018)}]{lee2018seed}
Grace~E. Lee and Aixin Sun. 2018.
\newblock \href {https://doi.org/10.1145/3209978.3209994} {Seed-driven document
  ranking for systematic reviews in evidence-based medicine}.
\newblock In \emph{The 41st International ACM SIGIR Conference on Research \&
  Development in Information Retrieval}, SIGIR ’18, page 455–464, New York,
  NY, USA. Association for Computing Machinery.

\bibitem[{Lee et~al.(2020)Lee, Yoon, Kim, Kim, Kim, So, and
  Kang}]{lee2020biobert}
Jinhyuk Lee, Wonjin Yoon, Sungdong Kim, Donghyeon Kim, Sunkyu Kim, Chan~Ho So,
  and Jaewoo Kang. 2020.
\newblock Biobert: a pre-trained biomedical language representation model for
  biomedical text mining.
\newblock \emph{Bioinformatics}, 36(4):1234--1240.

\bibitem[{Lehman et~al.(2019)Lehman, DeYoung, Barzilay, and
  Wallace}]{lehman-etal-2019-inferring}
Eric Lehman, Jay DeYoung, Regina Barzilay, and Byron~C. Wallace. 2019.
\newblock \href {https://doi.org/10.18653/v1/N19-1371} {Inferring which medical
  treatments work from reports of clinical trials}.
\newblock In \emph{Proceedings of the 2019 Conference of the North {A}merican
  Chapter of the Association for Computational Linguistics: Human Language
  Technologies, Volume 1 (Long and Short Papers)}, pages 3705--3717,
  Minneapolis, Minnesota. Association for Computational Linguistics.

\bibitem[{Maaten and Hinton(2008)}]{maaten2008visualizing}
Laurens van~der Maaten and Geoffrey Hinton. 2008.
\newblock Visualizing data using t-sne.
\newblock \emph{Journal of machine learning research}, 9(Nov):2579--2605.

\bibitem[{Marshall et~al.(2017)Marshall, Kuiper, Banner, and
  Wallace}]{marshall-etal-2017-automating}
Iain Marshall, Jo{\"e}l Kuiper, Edward Banner, and Byron~C. Wallace. 2017.
\newblock \href {https://www.aclweb.org/anthology/P17-4002} {Automating
  biomedical evidence synthesis: {R}obot{R}eviewer}.
\newblock In \emph{Proceedings of {ACL} 2017, System Demonstrations}, pages
  7--12, Vancouver, Canada. Association for Computational Linguistics.

\bibitem[{Mehra et~al.(2020)Mehra, Desai, Ruschitzka, and
  Patel}]{mehra2020hydroxychloroquine}
Mandeep~R Mehra, Sapan~S Desai, Frank Ruschitzka, and Amit~N Patel. 2020.
\newblock Hydroxychloroquine or chloroquine with or without a macrolide for
  treatment of covid-19: a multinational registry analysis.
\newblock \emph{The Lancet}.

\bibitem[{Minervini and Riedel(2018)}]{minervini2018adversarially}
Pasquale Minervini and Sebastian Riedel. 2018.
\newblock Adversarially regularising neural nli models to integrate logical
  background knowledge.
\newblock In \emph{Proceedings of the 22nd Conference on Computational Natural
  Language Learning}, pages 65--74.

\bibitem[{Morra et~al.(2018)Morra, Van~Thanh, Kamel, Ghazy, Altibi, Dat, Thy,
  Vuong, Mostafa, Ahmed et~al.}]{morra2018clinical}
Mostafa~Ebraheem Morra, Le~Van~Thanh, Mohamed~Gomaa Kamel, Ahmed~Abdelmotaleb
  Ghazy, Ahmed~MA Altibi, Lu~Minh Dat, Tran Ngoc~Xuan Thy, Nguyen~Lam Vuong,
  Mostafa~Reda Mostafa, Sarah~Ibrahim Ahmed, et~al. 2018.
\newblock Clinical outcomes of current medical approaches for middle east
  respiratory syndrome: A systematic review and meta-analysis.
\newblock \emph{Reviews in medical virology}, 28(3):e1977.

\bibitem[{Nye et~al.(2018)Nye, Li, Patel, Yang, Marshall, Nenkova, and
  Wallace}]{nye-etal-2018-corpus}
Benjamin Nye, Junyi~Jessy Li, Roma Patel, Yinfei Yang, Iain Marshall, Ani
  Nenkova, and Byron Wallace. 2018.
\newblock \href {https://doi.org/10.18653/v1/P18-1019} {A corpus with
  multi-level annotations of patients, interventions and outcomes to support
  language processing for medical literature}.
\newblock In \emph{Proceedings of the 56th Annual Meeting of the Association
  for Computational Linguistics (Volume 1: Long Papers)}, pages 197--207,
  Melbourne, Australia. Association for Computational Linguistics.

\bibitem[{Paszke et~al.(2019)Paszke, Gross, Massa, Lerer, Bradbury, Chanan,
  Killeen, Lin, Gimelshein, Antiga, Desmaison, Kopf, Yang, DeVito, Raison,
  Tejani, Chilamkurthy, Steiner, Fang, Bai, and Chintala}]{NEURIPS2019_9015}
Adam Paszke, Sam Gross, Francisco Massa, Adam Lerer, James Bradbury, Gregory
  Chanan, Trevor Killeen, Zeming Lin, Natalia Gimelshein, Luca Antiga, Alban
  Desmaison, Andreas Kopf, Edward Yang, Zachary DeVito, Martin Raison, Alykhan
  Tejani, Sasank Chilamkurthy, Benoit Steiner, Lu~Fang, Junjie Bai, and Soumith
  Chintala. 2019.
\newblock \href
  {http://papers.neurips.cc/paper/9015-pytorch-an-imperative-style-high-performance-deep-learning-library.pdf}
  {Pytorch: An imperative style, high-performance deep learning library}.
\newblock In \emph{Advances in Neural Information Processing Systems 32}, pages
  8024--8035. Curran Associates, Inc.

\bibitem[{Peymani et~al.(2016)Peymani, Ghavami, Yeganeh, Tabrizi, Sabour,
  Geramizadeh, Fattahi, Ahmadi, and Lankarani}]{peymani2016effect}
P~Peymani, S~Ghavami, B~Yeganeh, R~Tabrizi, S~Sabour, B~Geramizadeh,
  MR~Fattahi, SM~Ahmadi, and KB~Lankarani. 2016.
\newblock Effect of chloroquine on some clinical and biochemical parameters in
  non-response chronic hepatitis c virus infection patients: pilot clinical
  trial.
\newblock \emph{Acta bio-medica: Atenei Parmensis}, 87(1):46.

\bibitem[{Sackett(1997)}]{sackett1997evidence}
David~L Sackett. 1997.
\newblock Evidence-based medicine.
\newblock In \emph{Seminars in perinatology}, pages 3--5. Elsevier.

\bibitem[{Sheahan et~al.(2017)Sheahan, Sims, Graham, Menachery, Gralinski,
  Case, Leist, Pyrc, Feng, Trantcheva et~al.}]{sheahan2017broad}
Timothy~P Sheahan, Amy~C Sims, Rachel~L Graham, Vineet~D Menachery, Lisa~E
  Gralinski, James~B Case, Sarah~R Leist, Krzysztof Pyrc, Joy~Y Feng, Iva
  Trantcheva, et~al. 2017.
\newblock Broad-spectrum antiviral gs-5734 inhibits both epidemic and zoonotic
  coronaviruses.
\newblock \emph{Science translational medicine}, 9(396).

\bibitem[{Singh et~al.(2017)Singh, Marshall, Thomas, Shawe-Taylor, and
  Wallace}]{singh2017annotation}
Gaurav Singh, Iain~J. Marshall, James Thomas, John Shawe-Taylor, and Byron~C.
  Wallace. 2017.
\newblock \href {https://doi.org/10.1145/3132847.3132989} {A neural
  candidate-selector architecture for automatic structured clinical text
  annotation}.
\newblock In \emph{Proceedings of the 2017 ACM on Conference on Information and
  Knowledge Management}, CIKM ’17, page 1519–1528, New York, NY, USA.
  Association for Computing Machinery.

\bibitem[{Vaswani et~al.(2017)Vaswani, Shazeer, Parmar, Uszkoreit, Jones,
  Gomez, Kaiser, and Polosukhin}]{vaswani2017attention}
Ashish Vaswani, Noam Shazeer, Niki Parmar, Jakob Uszkoreit, Llion Jones,
  Aidan~N Gomez, \L~ukasz Kaiser, and Illia Polosukhin. 2017.
\newblock \href
  {http://papers.nips.cc/paper/7181-attention-is-all-you-need.pdf} {Attention
  is all you need}.
\newblock In I.~Guyon, U.~V. Luxburg, S.~Bengio, H.~Wallach, R.~Fergus,
  S.~Vishwanathan, and R.~Garnett, editors, \emph{Advances in Neural
  Information Processing Systems 30}, pages 5998--6008. Curran Associates, Inc.

\bibitem[{Wallace(2019)}]{ijcai2019-899}
Byron~C. Wallace. 2019.
\newblock \href {https://doi.org/10.24963/ijcai.2019/899} {What does the
  evidence say? models to help make sense of the biomedical literature}.
\newblock In \emph{Proceedings of the Twenty-Eighth International Joint
  Conference on Artificial Intelligence, {IJCAI-19}}, pages 6416--6420.
  International Joint Conferences on Artificial Intelligence Organization.

\bibitem[{Wallace et~al.(2016)Wallace, Kuiper, Sharma, Zhu, and
  Marshall}]{JMLR:v17:15-404}
Byron~C. Wallace, Jo{{\"e}}l Kuiper, Aakash Sharma, Mingxi~(Brian) Zhu, and
  Iain~J. Marshall. 2016.
\newblock \href {http://jmlr.org/papers/v17/15-404.html} {Extracting pico
  sentences from clinical trial reports using supervised distant supervision}.
\newblock \emph{Journal of Machine Learning Research}, 17(132):1--25.

\bibitem[{Wang et~al.(2019)Wang, Sun, and Xing}]{wang2019if}
Haohan Wang, Da~Sun, and Eric~P Xing. 2019.
\newblock What if we simply swap the two text fragments? a straightforward yet
  effective way to test the robustness of methods to confounding signals in
  nature language inference tasks.
\newblock In \emph{Proceedings of the AAAI Conference on Artificial
  Intelligence}, volume~33, pages 7136--7143.

\bibitem[{Wang et~al.(2020{\natexlab{a}})Wang, Lo, Chandrasekhar, Reas, Yang,
  Eide, Funk, Kinney, Liu, Merrill et~al.}]{wang2020cord}
Lucy~Lu Wang, Kyle Lo, Yoganand Chandrasekhar, Russell Reas, Jiangjiang Yang,
  Darrin Eide, Kathryn Funk, Rodney Kinney, Ziyang Liu, William Merrill, et~al.
  2020{\natexlab{a}}.
\newblock Cord-19: The covid-19 open research dataset.
\newblock \emph{arXiv preprint arXiv:2004.10706}.

\bibitem[{Wang et~al.(2020{\natexlab{b}})Wang, Zhang, Du, Du, Zhao, Jin, Fu,
  Gao, Cheng, Lu et~al.}]{wang2020remdesivir}
Yeming Wang, Dingyu Zhang, Guanhua Du, Ronghui Du, Jianping Zhao, Yang Jin,
  Shouzhi Fu, Ling Gao, Zhenshun Cheng, Qiaofa Lu, et~al. 2020{\natexlab{b}}.
\newblock Remdesivir in adults with severe covid-19: a randomised,
  double-blind, placebo-controlled, multicentre trial.
\newblock \emph{The Lancet}.

\bibitem[{WHO(2020)}]{Solidarity}
WHO. 2020.
\newblock \href
  {https://www.who.int/emergencies/diseases/novel-coronavirus-2019/global-research-on-novel-coronavirus-2019-ncov/solidarity-clinical-trial-for-covid-19-treatments}
  {Solidarity clinical trial for covid-19 treatments}.

\bibitem[{Wolf et~al.(2019)Wolf, Debut, Sanh, Chaumond, Delangue, Moi, Cistac,
  Rault, Louf, Funtowicz, and Brew}]{Wolf2019HuggingFacesTS}
Thomas Wolf, Lysandre Debut, Victor Sanh, Julien Chaumond, Clement Delangue,
  Anthony Moi, Pierric Cistac, Tim Rault, R'emi Louf, Morgan Funtowicz, and
  Jamie Brew. 2019.
\newblock Huggingface's transformers: State-of-the-art natural language
  processing.
\newblock \emph{ArXiv}, abs/1910.03771.

\bibitem[{Wong et~al.(2019)Wong, Siah, and Lo}]{wong2019estimation}
Chi~Heem Wong, Kien~Wei Siah, and Andrew~W Lo. 2019.
\newblock Estimation of clinical trial success rates and related parameters.
\newblock \emph{Biostatistics}, 20(2):273--286.

\bibitem[{Yang et~al.(2019)Yang, Xie, Lin, Li, Tan, Xiong, Li, and
  Lin}]{yang-etal-2019-end}
Wei Yang, Yuqing Xie, Aileen Lin, Xingyu Li, Luchen Tan, Kun Xiong, Ming Li,
  and Jimmy Lin. 2019.
\newblock \href {https://doi.org/10.18653/v1/N19-4013} {End-to-end open-domain
  question answering with {BERT}serini}.
\newblock In \emph{Proceedings of the 2019 Conference of the North {A}merican
  Chapter of the Association for Computational Linguistics (Demonstrations)},
  pages 72--77, Minneapolis, Minnesota. Association for Computational
  Linguistics.

\bibitem[{Zhang et~al.(2020)Zhang, Yu, Mei, Tang, Zhang, and
  Li}]{zhang2020unlocking}
Tengteng Zhang, Yiqin Yu, Jing Mei, Zefang Tang, Xiang Zhang, and Shaochun Li.
  2020.
\newblock Unlocking the power of deep pico extraction: Step-wise medical ner
  identification.
\newblock \emph{arXiv preprint arXiv:2005.06601}.

\end{thebibliography}
\bibliographystyle{acl_natbib}

\end{document}